\documentclass[letterpaper, 10 pt, conference]{ieeeconf}  

\IEEEoverridecommandlockouts                              
\usepackage[style=ieee, sorting=none, citestyle=numeric-comp]{biblatex}

\usepackage[%
    colorlinks=True,
    citecolor=blue,
    pdfborder={0 0 0},
    linkcolor=red
]{hyperref}

\usepackage{algorithm} 
\usepackage{algpseudocode}

\usepackage{algorithm}
\usepackage{algorithmicx}
\usepackage{algpseudocode}
\usepackage{amsmath}

\usepackage[utf8]{inputenc}
\usepackage[T1]{fontenc}

\usepackage{graphicx}
\graphicspath{ {./Figures/} }

\usepackage{graphics} 
\usepackage{epsfig} 
\usepackage{mathptmx} 
\usepackage{times} 
\usepackage{amsmath} 

\usepackage{amssymb}  
\usepackage{bm}

\title{\LARGE \bf
Real-Time Bayesian Detection of Drift-Evasive GNSS Spoofing in Reinforcement Learning Based UAV Deconfliction
}

\author{Deepak Kumar Panda$^{1}$ and Weisi Guo$^{1}$
\thanks{*This work was supported by the Royal Academy of Engineering and the
Office of the Chief Science Adviser for National Security under the UK
Intelligence Community Postdoctoral Research Fellowship programme}
\thanks{$^{1}$The authors are with Centre for Connected and Assured Autonomy, Faculty of Engineering and Applied Sciences, Cranfield University MK43 0AL. 
{\tt\small Deepak.Panda@cranfield.ac.uk, Weisi.Guo@cranfield.ac.uk}}%
}

\begin{document}

\maketitle
\thispagestyle{empty}
\pagestyle{empty}

\begin{abstract}
Autonomous unmanned aerial vehicles (UAVs) rely on global navigation satellite system (GNSS) pseudorange measurements for accurate real-time localization and navigation. However, this dependence exposes them to sophisticated spoofing threats, where adversaries manipulate pseudoranges to deceive UAV receivers. Among these, drift-evasive spoofing attacks subtly perturb measurements, gradually diverting the UAV’s trajectory without triggering conventional signal-level anti-spoofing mechanisms. Traditional distributional shift detection techniques often require accumulating a threshold number of samples, causing delays that impede rapid detection and timely response. Consequently, robust temporal-scale detection methods are essential to identify attack onset and enable contingency planning with alternative sensing modalities, improving resilience against stealthy adversarial manipulations. This study explores a Bayesian online change point detection (BOCPD) approach that monitors temporal shifts in value estimates from a reinforcement learning (RL) critic network to detect subtle behavioural deviations in UAV navigation. Experimental results show that this temporal value-based framework outperforms conventional GNSS spoofing detectors, temporal semi-supervised learning frameworks, and the Page-Hinkley test, achieving higher detection accuracy and lower false-positive and false-negative rates for drift-evasive spoofing attacks.
\end{abstract}

\section{INTRODUCTION}
Unmanned aerial vehicles (UAVs) are increasingly being deployed across diverse civilian applications, including environmental monitoring and surveillance \cite{liu2022uav}, last-mile logistics \cite{shao2025efficient}, and precision agriculture \cite{tokekar2016sensor}. Their capacity to operate autonomously within complex, dynamic, and high-risk environments enables them to be deployed in sensitive and safety-critical missions. UAV operational autonomy is primarily driven by the global navigation satellite system (GNSS), which provides continuous real-time positioning and velocity estimation, and hence it plays an essential role in enabling essential navigation functions, including trajectory planning, geo-fencing, adherence to designated flight corridors, and automated return-to-home procedures \cite{zhang2018intelligent}.

However, the unencrypted civilian GNSS signals expose the UAVs to malicious exploitation, as they can be spoofed using counterfeit signals 
thus manipulating their position, velocity, and time (PVT) estimates \cite{ioannides2016known}. Successful spoofing can cause UAVs to deviate from intended paths, enter restricted airspace, or be hijacked \cite{noh2019tractor}.  As UAVs increasingly integrate reinforcement learning (RL) for decision-making and trajectory planning \cite{wang2019autonomous}, their dependence on trustworthy GNSS inputs becomes even more critical. A compromised GNSS feed can mislead, not only the navigation system but also the underlying policy optimization loop of the RL agent, resulting in the navigation mission compromise. To mitigate such risks, a wide range of GNSS spoofing detection strategies have emerged. Traditional signal-level defences identify anomalies in Doppler shift, correlation distortion, signal strength fluctuations or power spectral density \cite{broumandan2017approach, ioannides2016known}. Sensor-fusion approaches provide improvement over signal anomaly detection techniques, which augment GNSS with inertial measurement units (IMUs) or odometry-based checks to flag inconsistencies \cite{meng2025trusted}. More recently, machine learning-based methods have gained popularity, especially supervised approaches that use labelled datasets to train classifiers for spoofing detection \cite{borhani2020deep, iqbal2024deep}. However, supervised learning is constrained by a critical limitation: the evolving nature of spoofing attacks. Drift-based spoofing strategies are inherently adaptive, often tailored to specific mission contexts or defence bypasses, making it impractical to obtain representative and comprehensive labelled datasets in advance.

To address this, unsupervised anomaly detection frameworks have been proposed, particularly in settings where well-labelled datasets are unavailable. These methods aim to learn the normal operating distribution of the UAV—either through generative models such as variational autoencoders \cite{panda2025generative}, by projecting data into a minimal hypersphere \cite{ruff2018deep}, or by employing contrastive loss functions to distinguish in-distribution from out-of-distribution samples \cite{yang2022class}. While these approaches can effectively detect anomalies in an unsupervised manner, they typically require the collection of a sufficient number of samples to establish a reference distribution or embedding metric. In UAV navigation, however, timely detection is critical, and there may be limited opportunity to accumulate observations before making decisions. Therefore, detection methods must operate online and continuously monitor temporal sequences to promptly identify potential attacks, especially evasive attacks that gradually incorporate the perturbations into the UAV sensors.

Existing unsupervised temporal anomaly detection methods commonly rely on modeling sequential dependencies through recurrent autoencoders \cite{yang2020feedback} or temporal convolutional networks \cite{li2022unsupervised}, which estimate predictive intervals over time. While such models capture temporal patterns, they are prone to high sensitivity to non-adversarial distributional shifts and often require specialized training pipelines. 

In this paper, we present a novel detection framework that leverages the temporal sequence of state-action-value (Q-value) estimates from a trained RL critic network as a direct indicator of decision confidence under nominal scenarios. We show that monitoring these Q-values enables early detection of drift-evasive GNSS spoofing attacks that conventional observation-based detectors fail to identify. Specifically, we employ Bayesian online change point detection (BOCPD) \cite{adams2007bayesian} to infer a probabilistic posterior over the run lengths of the Q-value sequence, providing timely detection of distributional shifts with principled uncertainty quantification. Hence, the main contributions of the paper can be summarized as follows:
\begin{itemize}
    \item \textbf{BOCPD for Drift-Evasive Spoofing Attack from Latent $Q$ values:} Utilizing the BOCPD with the latent q-values for fast spoofing attack detection from the temporal state-action value estimates. 
    \item \textbf{Validation with Semi-Supervised and Statistical Anomaly Detectors: } Empirically validating that the approach with BOCPD with respect to semi-supervised time-series baselines \cite{malhotra2015long}, conventional signal-level methods \cite{ioannides2016known}, and the Page-Hinkley test \cite{montiel2018scikit} in both detection accuracy, false-positive and false-negative rates.
\end{itemize}
\section{Overall Methodology and GNSS Signal Description}
\begin{figure*}[thpb]
      \centering
      \includegraphics[scale=0.40]{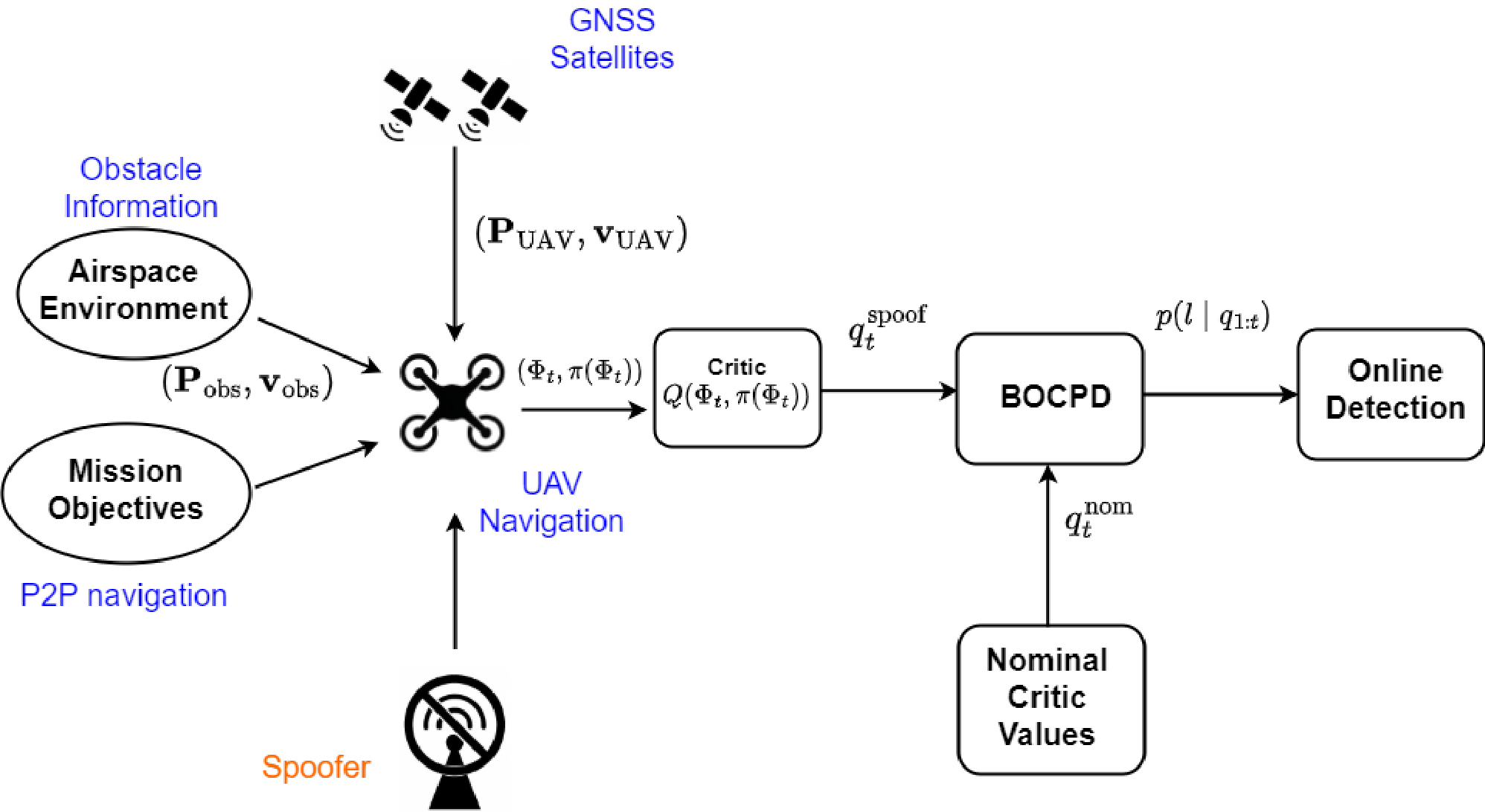}
      \caption{Schematic of the GNSS drift spoof detection mechanism from the latent q-values.}
      \label{figure1_schematic}
\end{figure*}
As shown in Figure \ref{figure1_schematic}, a point-to-point UAV navigation task is considered where it is required to reach a designated goal location $\mathbf{P}_g$ while avoiding a dynamic 3D obstacle within the operational airspace. At each time step, the UAV obtains observations comprising its position and velocity $\left (\mathbf{P}_{\text{UAV}}, \mathbf{v}_{\text{UAV}} \right )$ along with the position and velocity of the obstacle $\left (\mathbf{P}_{\text{obs}}, \mathbf{v}_{\text{obs}} \right )$. The position estimate $ \mathbf{P}_{\text{UAV}}$ is derived from the GNSS signals via satellite-based triangulation. Hence, its position, velocity, mission characteristics and obstacle info are aggregated into a state vector $\Phi$ which serves as an input to the control policy. A deep deterministic policy gradient (DDPG) is used to train the policy network, $\pi$, which maps the states $\Phi$ to action $\pi \left (\Phi \right )$ from the critic network $Q \left ( \cdot \right )$.

In order to model adversarial conditions, a GNSS spoofing attack is introduced in which the UAV’s estimated position is progressively displaced from its true trajectory, rendering conventional anomaly detection methods ineffective. To detect such drift-evasive attacks, we employ a Bayesian online change point detection (BOCPD) algorithm that continuously monitors the temporal sequence of value estimates from the critic network. Specifically, BOCPD tracks deviations of the spoofed critic signal  $q^{\text{spoof}}_t$ relative to the nominal critic distribution values, $q^{\text{nom}}_t$.  By recursively updating the posterior distribution over run lengths, $p \left ( l \mid q_{1:t} \right )$, the algorithm enables timely identification of distributional shifts induced by the attack.
\subsection{GNSS Signal Description and Position Estimation}
In order to estimate the position of the UAV along with the receiver clock bias, an iterative least-squares approach is adopted using the pseudorange measurements from multiple pseudo-satellites \cite{psiaki2016gnss}. If we consider $N$ satellites located at position $\mathbf{S}_i \in \mathbb{R}^3$ where $i = 1, \cdots, N$. The measured pseudorange to the $i^{\text{th}}$ satellite is denoted by $\kappa_i$. Hence, each pseudorange measurement is expressed as,
\begin{equation}\label{eq:1}
    \kappa_{i}=\left\|\mathbf{P}-\mathbf{S}_{i}\right\|+b+\varepsilon_{i},
\end{equation}
where $\mathbf{P} \in \mathbb{R}^3$ represent the unknown receiver position, $b \in \mathbb{R}$ is the receiver clock bias relative to the satellite time base and $\epsilon_i$ is the measurement noise modelled using a zero-mean Gaussian noise. Hence, the state vector $\mathbf{P} = \begin{bmatrix}
X & Y & Z & b \\
\end{bmatrix} $ are to be estimated. At each iteration $k$, given a current estimate $P^{\left (k \right )}$, the following steps are performed to obtain the final receiver position.
\begin{itemize}
    \item For each satellite $i$, the model-predicted pseudorange $\hat{\kappa}_i$ is calculated as:
    \begin{equation} \label{eq:2}
    \hat{\kappa}_{i}=\left\|\mathbf{P}^{(k)}-\mathbf{S}_{i}\right\|+b^{(k)}.
    \end{equation}
    \item The residual is computed which quantifies the discrepancy between the measured and predicted pseudoranges which is given as:
    \begin{equation} \label{eq:3}
        \delta_{i}=\kappa_{i}-\hat{\kappa}_{i}=\kappa_{i}-\left\|\mathbf{P}^{(k)}-\mathbf{S}_{i}\right\|-b^{(k)} .
    \end{equation}
    \item The Jacobian matrix $\mathbf{H} \in \mathbb{R}^{N \times 4}$ captures the sensitivity of each residual to changes in the state vector, where each row is defined as $\mathbf{H}_{i}=\left[\frac{\partial \hat{\kappa}_{i}}{\partial X}, \frac{\partial \hat{\kappa}_{i}}{\partial Y}, \frac{\partial \hat{\kappa}_{i}}{\partial Z}, \frac{\partial \hat{\kappa}_{i}}{\partial b}\right]$. Thus, we can write the derivatives as:
    \begin{equation} \label{eq:4}
         \frac{\partial \hat{\kappa}_{i}}{\partial \mathbf{P}}=\frac{\mathbf{P}^{(k)}-\mathbf{S}_{i}}{\left\|\mathbf{P}^{(k)}-\mathbf{S}_{i}\right\|}, \quad \frac{\partial \hat{\kappa}_{i}}{\partial b}=1.
    \end{equation}
    Hence, the Jacobian row corresponding to the satellite $i$ is:
    \begin{equation} \label{eq:5}
    \mathbf{H}_{i}=\left[\begin{array}{llll}
    \frac{X^{(k)}-S_{i}^{x}}{\left\|\mathbf{P}^{(k)}-\mathbf{S}_{i}\right\|} & \frac{Y^{(k)}-S_{i}^{y}}{\left\|\mathbf{P}^{(k)}-\mathbf{S}_{i}\right\|} & \frac{Z^{(k)}-S_{i}^{z}}{\left\|\mathbf{P}^{(k)}-\mathbf{S}_{i}\right\|} & 1
    \end{array}\right].
    \end{equation}
    \item The correction to the estimated position vector is obtained via the least-square solution given as,
    \begin{equation} \label{eq:6}
        \Delta \mathbf{P}=\left(\mathbf{H}^{\top} \mathbf{H}\right)^{-1} \mathbf{H}^{\top} \boldsymbol{\delta}. 
    \end{equation}
    The updated state estimate is given as:
    \begin{equation} \label{eq:7}
        \mathbf{P}^{\left ( k+1 \right )}=\mathbf{P}^{\left ( k \right )}+\Delta \mathbf{P}.
    \end{equation}
\end{itemize}
\subsection{Transition Probability Model $\mathcal{T}$}
The transition dynamics are defined by a hybrid kinematic and reactive flow-field-based model as defined in \cite{panda2025curriculum}. Given the current UAV position $\mathbf{x}_t$ and action $\mathbf{a}_t$, the next position is computed using the following relation:
\begin{equation} \label{eq:8}
    \mathbf{P}_{t+1} = \mathbf{P}_t + \Delta t \cdot \bar{u}_t,
\end{equation}
where $\bar{u}_t$ is the resultant motion vector given as, $\bar{u}_t = \left [\mathbf{I} + \mathbf{M}_{\text{rep}} + \mathbf{M}_{\text{tan}} \right ] \left (\mathbf{u}_t - \mathbf{v}_{\text{obs}} \right ) + \mathbf{v}_{\text{obs}} $ as defined in \cite{panda2025curriculum}. Here $\mathbf{u}_t$ represents the initial attractive vector pointing from current UAV position to the goal, $\mathbf{M}_{\text{rep}}$ and $\mathbf{M}_{\text{tan}}$ are repulsive and tangential potential field matrices respectively and $\mathbf{v}_{\text{obs}}$ is the velocity of the nearest dynamic obstacle. The transition in (\ref{eq:8}) represents a nonlinear, time-varying model governed by both control input and obstacle dynamics. 
\subsection{State Space}
The state-vector at time $t$, $\Phi_t \in \mathbb{R}^3$ incorporates relative geometry and dynamic information given as follows:
\begin{equation} \label{eq:9}
{\Phi}_t = \left[ 
\frac{P_{\text{rel}} \cdot \left( P_{\text{rel}} - \mathfrak{r}_{\text{obs}} \right)}{\| P_{\text{rel}}\|},\ 
\mathbf{P}_{\text{goal}} - \mathbf{P}_{\text{UAV}},\ 
\mathbf{v}_{\text{obs}} 
\right],
\end{equation}
where $P_{\text{rel}} = \mathbf{P}_{\text{obs}} - \mathbf{P}_{\text{UAV}}$. Here, $\mathbf{P}_{\text{UAV}} \in \mathbb{R}^3 $ represents the UAV position as calculated from the GNSS pseudomeasurements as given in (\ref{eq:7}), $\mathbf{P}_{\text{obs}} \in \mathbb{R}^3 $ represents the position of the dynamic obstacle at time $t$ obtained via ADS-B, $\mathbf{P}_{\text{goal}} \in \mathbb{R}^3 $ represents the mission goal location, $\mathbf{v}_{\text{obs}}$ represents the velocity of the obstacle and $\mathfrak{r}_{\text{obs}}$ represents the obstacle radius. 
\subsection{Action Space}
The agent outputs a continuous action vector $\mathbf{a}_t \in \mathbb{R}^3$ representing the navigation control parameters $\mathbf{a}_t =  \left[ \varrho_0, \varsigma_0, \theta \right ]$,where $\varrho_0$ and $\varsigma_0$ represents the strength of the repulsive and tangential potential field and $\theta$ represents the tangential vector rotation angle. The parameters shape the resultant control vector $\bar{\mathbf{u}}$ by modulating the influence of the obstacle repulsion and goal-oriented attractive force. 
\subsection{Reward Function}
The reward function encourages the obstacle avoidance and efficient goal-oriented motion, which is defined as follows:
\begin{itemize}
    \item \textbf{Penalty for collision:}
    \begin{equation} \label{eq:10}
    r^{\text{coll}}_{t}=-1+\frac{d-\mathfrak{r}_{\mathrm{obs}}}{\mathfrak{r}_{\mathrm{obs}}}, \quad \text { if } d \leq \mathfrak{r}_{\mathrm{obs}}
    \end{equation}
    \item \textbf{Penalty for being in the threat zone near the obstacle:}
    \begin{equation} \label{eq:11}
     r^{\text{thr}}_{t}=-0.3+\frac{d-\left(\mathfrak{r}_{\mathrm{obs}}+\xi\right)}{\mathfrak{r}_{\mathrm{obs}}+\xi}, \quad \text { if } \mathfrak{r}_{\mathrm{obs}}<d<\mathfrak{r}_{\mathrm{obs}}+\xi.
    \end{equation}
    with $\xi = 0.4$m being the safety margin.
    \item \textbf{Goal Seeking Reward:}
    \begin{equation} \label{eq:12}
    r^{\text{goal}}_{t}=-\frac{\left\|\mathbf{P}_{\text {goal }}-\mathbf{P}_{t+1}\right\|}{\left\|\mathbf{P}_{\text {goal }}-\mathbf{P}_{0}\right\|}+\mathbb{I}_{\text {goal }} \cdot 3
    \end{equation}.
    where $\mathbb{I}_{\text {goal }} = 1$ if the UAV is within the success threshold of the goal.
\end{itemize}
\section{Drift Evasive GNSS Spoofing Attack}
Autonomous UAVs that rely on GNSS for navigation are vulnerable to spoofing attacks \cite{ioannides2016known}, with a drift-evasive attack where the estimated position is gradually manipulated to avoid abrupt deviations detectable by residual-signal based monitoring. If the true position of the UAV at time $t$ is $\mathbf{P}_t$ and the attacker's target position is $\mathbf{P}^{\text{tar}}_t$, thus the spoofed position evolves as: 
\begin{equation} \label{eq:13}
    \mathbf{P}^{\text{spoof}}_t = \left (1- \alpha_t \right ) \cdot \mathbf{P}_t + \alpha_t \cdot \mathbf{P}^{\text{tar}}_t,
\end{equation}
where, $\alpha_t = \min \left (1, \frac{t-t_{\text{start}}}{T_{\text{drift}}} \right )$. The smooth interpolation as shown in (\ref{eq:13}) maintains inertial consistency and reduces the chances of flagging the anomaly detectors. Hence, the attacker spoofs the UAV receivers with the pseudorange values given as:
\begin{equation} \label{eq:14}    \hat{\kappa}^{\text{spoof}}_{i}=\left\|\mathbf{P}^{\text{spoof}}_t-\mathbf{S}_{i}\right\|+b^{(k)} .
\end{equation}
The following assumptions are considered for the adversary to realistically inject the drift evasive attack.
\begin{itemize}
    \item \textbf{Timing synchronization:} The attacker is physically close to the UAV to maintain realistic signal delays.
    \item \textbf{Controlled signal power:} Transmissions slightly exceed authentic signals without triggering power alarms.
    \item \textbf{Direction of arrival emulation:} Multiple transmission points prevent detection of anomalous signal origins.
    \item \textbf{Alignment with drift path:} The spoofed pseudoranges remain consistent with satellite geometry to avoid kinematic inconsistencies.
\end{itemize}
\section{Bayesian Online Change Detection}
In the presence of stealthy GNSS spoofing, conventional threshold-based and statistical anomaly detectors often fail to identify gradual drifts in state estimates. To mitigate this, we employ a BOCPD framework \cite{adams2007bayesian}, which monitors the values $Q \left (\Phi, a \right )$ from the critic network. Let $\left \{q_1, \cdots, q_t \right \}$ be time-series of values $Q \left (\Phi^{\text{spoof}}_t, a^{\text{spoof}}_t \right )$, where $\Phi^{\text{spoof}}_t$ represents the potentially spoofed state at time $t$. BOCPD recursively infers the posterior distribution over the run length $l_t$, which denotes the time since the most recent change point:
\begin{equation} \label{eq:15}
     l_{t}:=\left\{\begin{array}{ll}
0, & \text { if } t \text { is a change point } \\
l_{t-1}+1, & \text { otherwise }
\end{array}\right.
\end{equation}
The goal is to compute the joint posterior over the run-length and observed data.
\begin{equation} \label{eq:16}
    P\left(l_{t}, q_{1: t}\right)=\sum_{l_{t-1}} P\left(l_{t} \mid l_{t-1}\right) \cdot P\left(q_{t} \mid q_{t-l_{t}: t-1}\right) \cdot P\left(l_{t-1}, q_{1: t-1}\right).
\end{equation}
The recursion involves:
\begin{itemize}
    \item \textbf{Hazard function:} Probability that a change occurs at time $t$ given a run length $l$, as $P\left(r_{t}=0 \mid r_{t-1}\right)=H\left(r_{t-1}\right)$.
    \item \textbf{Predictive Model $P\left(q_{t} \mid q_{t-l_{}: t-1}\right)$} : The likelihood of the new data under a probabilistic model of the current segment.
\end{itemize}
The  value estimates from the critic network are assumed to follow a Gaussian model, $q_t \sim \mathcal{N} \left (\mu_t, \sigma^2 \right )$. The mean and the standard deviation is computed from the critic value estimates obtained from nominal trajectory data $\mu_0 = \mathbb{E} \left [q^{\text{nom}}_t \right ]$, and the standard deviation $\sigma_0 = \text{Var} \left [q^{\text{nom}}_t \right ]$. The recursive update formula for the mean is given as follows:
\begin{equation} \label{eq:17}
    \mu_{l, t}=\frac{l \cdot \mu_{l-1, t-1}+q_{t}}{l+1}.
\end{equation}
Given the Gaussian assumption and the conjugacy of the Normal-inverse-gamma prior, the predictive probability is given as,
\begin{equation} \label{eq:18}
    P\left(q_{t} \mid \mu_{l, t-1}, \sigma^{2}\right)=\frac{1}{\sqrt{2 \pi \sigma^2_{\text{overall}}}} \exp \left(-\frac{\left(q_{t}-\mu_{l, t-1}\right)^{2}}{2\sigma^2_{\text{overall}}}\right)
\end{equation}
 where, $\sigma^2_{\text{overall}} =  \sigma^{2}+\sigma_{l, t-1}^{2}$. After normalization the posterior run length distribution $P \left ( l_t \mid q_{1:t} \right )$ is obtained from which the most probable run length $\hat{l} =  \arg \max_l L_{t,l} $ is used to signal the deviation due to spoofing attack. The overall process is presented in Algorithm \ref{alg:BOCPD_q}.
\begin{algorithm}[h] 
    \caption{Online Bayesian Changepoint Detection for Drift Evasive Spoofing Attack}
    \label{alg:BOCPD_q}
    \noindent\begin{minipage}{\linewidth}
    \begin{algorithmic}[1]  
    \Require Trained DDPG policy network $\pi$, DDP critic network $Q$. 
    \State \textbf{Initialize} $t \leftarrow 1$, set initial run-length $l_0 = 0$, prior critic mean $\mu_0$ and variance $\sigma^2$ from value estimate from nominal trajectories $q^{\text{nom}}_t$.
    \While{UAV mission not complete}
        \State Obtain GNSS pseudoranges $\kappa_i$, and estimate the UAV receiver position as per (\ref{eq:7}).
        \State Using the computed position and velocity and obstacle information $\left (\mathbf{P}_{\text{obs}}, \mathbf{v}_{\text{obs}} \right )$, formulate the observation $\Phi_t$.
        \State Based on the observation $\Phi_t$, take the action $a_t = \pi \left (\Phi_t \right )$.
        \State From the critic, obtain the real-time value estimate $q_t = Q \left (\Phi_t, a_t \right )$.
        \State Compute predictive probability $p\left  (q_t | \mu_t, \sigma^2 \right )$ using Gaussian likelihood as per (\ref{eq:16}).
        \State Update run-length posterior for $l_t$ as per the following relations:
        \begin{equation} \label{eq:19}
        \begin{aligned}
        \quad P\left (l_t = l+1 \right ) \propto P\left (l_{t-1} = l \right ) \cdot \left (1 - H \right) \cdot p \left (q_t \mid \mu_{l,t-1}, \sigma^2 \right ) \\
        \quad P\left (l_t = 0 \right ) \propto \sum_{l} P\left (l_{t-1} = l \right ) \cdot H \cdot p \left (q_t \mid \mu_{l,t-1}, \sigma^2 \right)
        \end{aligned}
        \end{equation}
        where $H$ is the constant hazard rate.
        \State Normalize posterior: $\sum_{l} P \left (l_t = l \right ) = 1$.
        \State Update sufficient statistics $\mu_{l,t}$ for each possible run-length $l$ using the recursive relation as per (\ref{eq:17}).
        \State Record $\hat{l}_t = \arg\max_l P\left (l_t = l \right )$ as the most probable run-length estimate.
        \State If $\hat{l}_t \leq \tau$ then flag GNSS spoofing (change point likely occurred).
        \State $t \leftarrow t+1$
    \EndWhile
\end{algorithmic}
\end{minipage}
\end{algorithm}
\section{Results and Discussions}
\subsection{Numerical Implementation}
A DDPG agent for UAV navigation was trained on GNSS-derived position estimates under realistic conditions, including sensor noise and dynamic 3D obstacles as per \cite{panda2025curriculum} and GNSS uncertainty. The UAV was initialized from random positions within bounded airspace and operated under kinematic constraints on yaw, climb rate, and pitch angles to ensure feasible manoeuvres. GNSS pseudoranges were simulated with signal geometry and clock errors, and positions were estimated as per (\ref{eq:7}). The state-space was formulated with the estimated position and velocity, along with the information from the dynamic 3D obstacles, similar to the trajectory in \cite{panda2025curriculum}.

The agent training spanned 200 episodes, with a maximum of 500 steps per episode. It was trained with a replay buffer of 1e6 transitions, a discount factor of 0.99, learning rates of 0.001 for both actor and critic networks, and exploration noise scale linear decreasing from 0.3 to 0.1. A drift-evasive attack was applied, after $t=100 s$ which interpolates UAV's estimated position towards $\left (0,0,0 \right )$ over 50 time-steps to avoid abrupt anomaly deviation. Spoofed signals matched legitimate GNSS characteristics in amplitude, phase, and delay, evading conventional defences such as RPM, correlation distortion, NMA, and DOA analysis as listed in \cite{ioannides2016known}. To detect these subtle position manipulations, the BOCPD framework monitors critic value estimates $q_t$ from the obtained temporal state and action pairs. Change point likelihoods were estimated using Gaussian predictive updates and a hazard function, with a run-length distribution matrix tracking shifts in decision dynamics to flag anomalies in real time.
\subsection{Training Results and Distribution Shift due to GNSS Spoofing Attack}
\begin{figure}[thpb]
      \centering
      \includegraphics[scale=0.18]{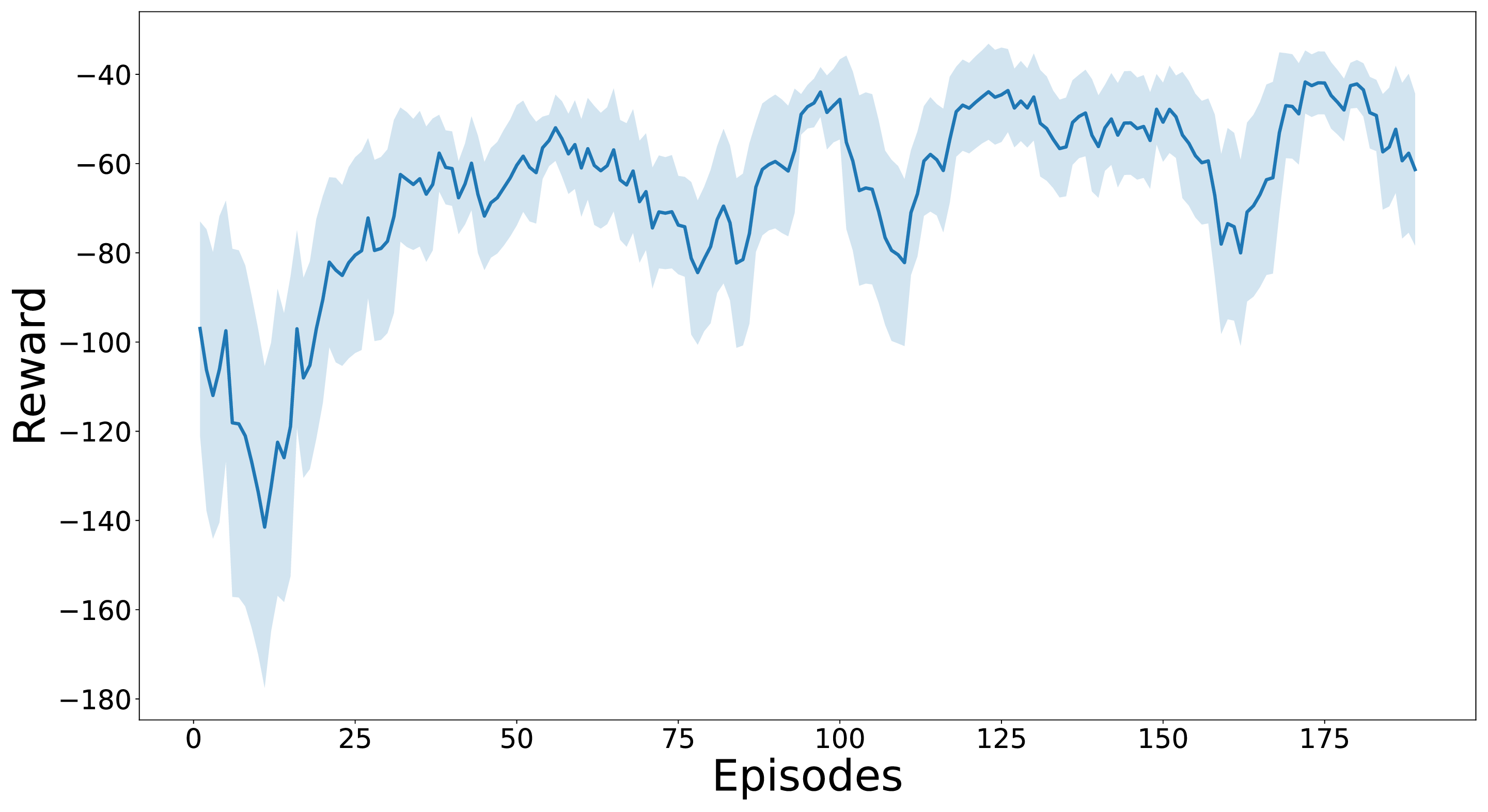}
      \caption{The rolling window mean and standard deviation of the episodic reward for training DDPG.}
    \label{figure2_training_convergence}
\end{figure}

\begin{figure}[thpb]
      \centering
      \includegraphics[scale=0.18]{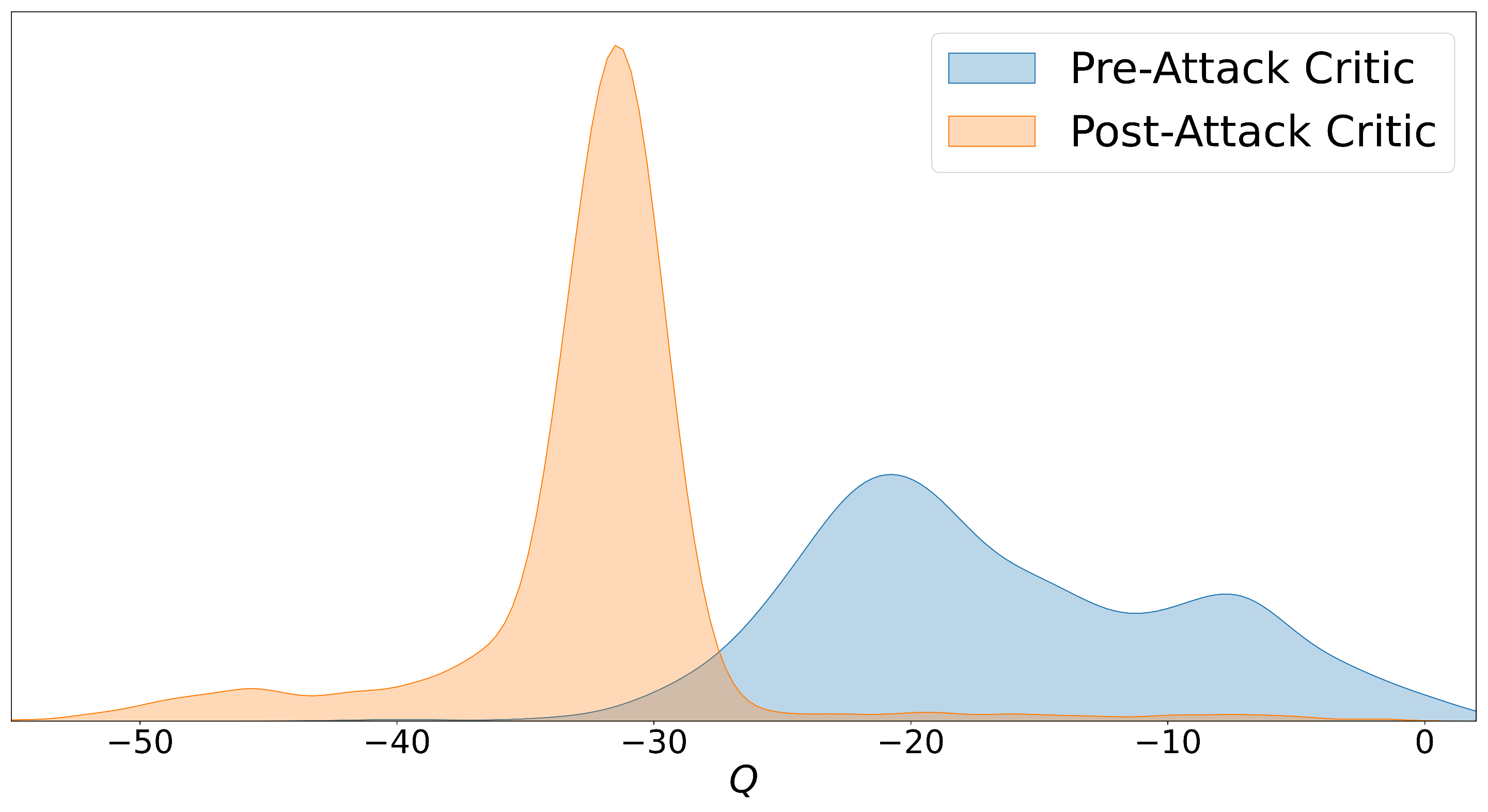}
      \caption{The comparison of the state-action value estimates $q_t$ on a sample space while considering pre- and post-attack UAV trajectories.}
      \label{figure3_distribution_shift}
\end{figure}

\begin{figure}[thpb]
      \centering
      \includegraphics[scale=0.18]{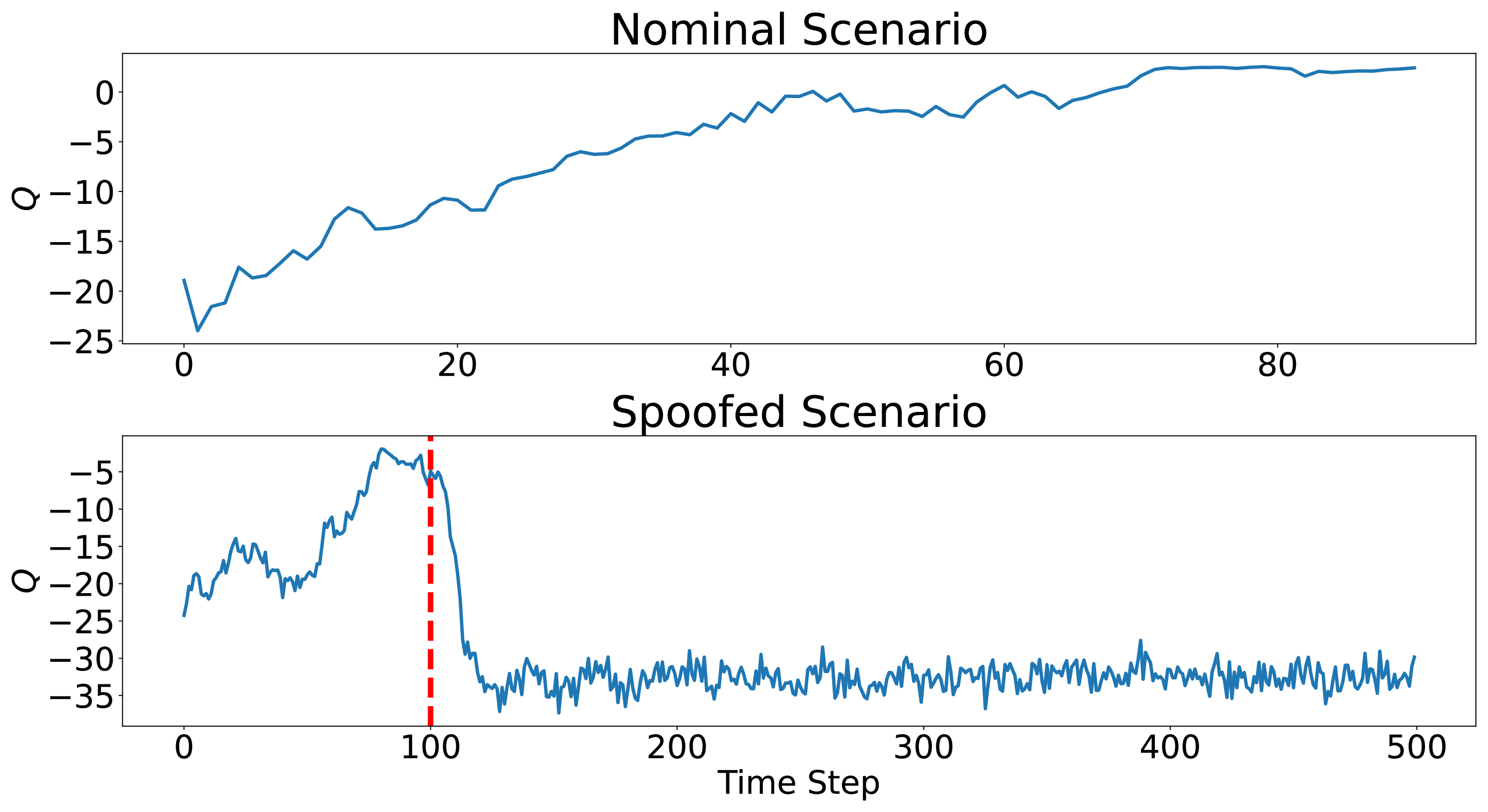}
      \caption{The comparison of the state-action value estimates $q_t$ on a temporal space for the nominal trajectories and the trajectory before and after the GNSS spoofing attack.}
      \label{figure4_temporal_shift}
\end{figure}

As illustrated in Figure \ref{figure2_training_convergence}, the training performance of the DDPG agent is evaluated over 200 episodes. The episodic rewards were smoothed using a moving-average filter with a window length of 10,  to visualize the learning trends of the agent. During the initial 30 episodes, the agent explores the action space randomly to enhance state-action coverage and promote diversity of experiences, which is critical for effective policy learning in subsequent episodes. Following the exploration phase, a consistent improvement in performance is observed, as indicated by the upward trend in the smoothed reward trajectory. The learning process exhibits gradual convergence, with the agent achieving a relatively stable performance within the 200-episode training horizon, as evident in Figure \ref{figure2_training_convergence}.
As illustrated in Figure \ref{figure3_distribution_shift}, the distribution of state-action value estimates demonstrates a clear shift following the GNSS spoofing attack initiated at $t=100s$. The post-attack distribution exhibits a significant leftward displacement, indicating a reduction in the agent's trust in state-action evaluations due to corrupted GNSS observations. In the temporal space, as shown in Figure \ref{figure4_temporal_shift}, the nominal scenario shows gradual convergence of the critic values towards zero, signifying successful mission completion. In contrast, under spoofed conditions, the critic value estimates remain persistently low after $t=100s$, failing to converge and ultimately resulting in mission compromise and emergency UAV landing. Detecting such abrupt changes in value estimates is therefore crucial for timely identifying the stealthy GNSS spoofing attacks, facilitating swift transition to alternative sensing modalities or fail-safe navigation strategies, thereby preserving operational integrity.
\subsection{Detection Results}
\begin{figure}[thpb]
      \centering
      \includegraphics[scale=0.18]{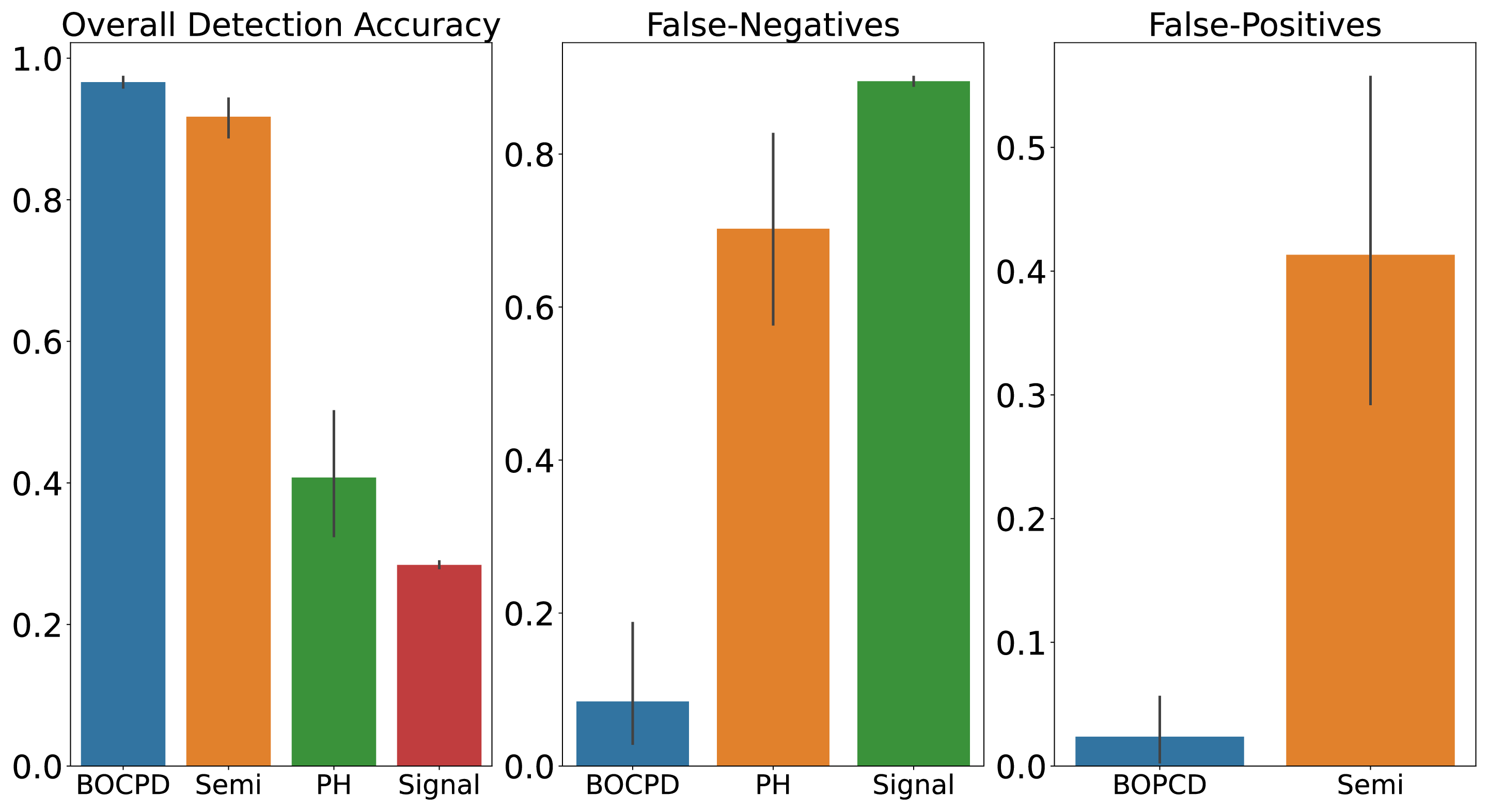}
      \caption{Comparison of GNSS spoofing detection performance over 20 test episodes. Shown are overall accuracy (left), false-negative rates (center), and false-positive rates (right) for BOCPD, Semi-supervised LSTM Autoencoder, Page-Hinkley (PH), and signal thresholding methods. Error bars indicate standard deviation across episodes.}
      \label{figure5_detection_metrics}
\end{figure}

Figure \ref{figure5_detection_metrics} provides a comparative evaluation of the proposed BOCPD against baseline methods, including semi-supervised LSTM, the Page-Hinkley (PH) statistic, and conventional signal thresholding for GNSS spoofing detection. BOCPD consistently demonstrates superior overall detection accuracy, approaching near-perfect rates, whereas PH and signal-based methods show significantly lower accuracy due to delayed detection or inconsistent identification of attack onset $t>100s$. Correspondingly, false-negative analysis highlights frequent missed detections by PH and signal-based approaches, while the semi-supervised method exhibits elevated false-positive rates due to its sensitivity to environmental deviations. In contrast, BOCPD maintains a robust balance, effectively minimizing both false negatives and false positives, thereby confirming its efficacy and reliability for timely and precise spoofing detection in adversarial GNSS conditions.
\section{Conclusion}
In this paper, a study was conducted to demonstrate BOCPD,  applied to RL value estimates, provides an effective and robust method for detecting drift-evasive GNSS spoofing attacks in autonomous UAV navigation. By continuously monitoring latent state-action-value sequences, BOCPD effectively identifies subtle yet critical shifts in behavioral characteristics from the latest value estimates typically overlooked by traditional signal-level and sequential detection methods. The UAV position estimation was performed using an iterative least-squares approach, utilizing pseudorange measurements from multiple GNSS satellites. The spoofing was conducted using manipulated pseudorange measurements to manipulate the position estimation towards targeted location in the airspace. Experimental evaluations confirmed that the BOCPD-based framework achieves superior detection accuracy, significantly reducing false positives and false negatives relative to baseline methods such as semi-supervised learning and Page-Hinkley tests. Future works involve integrating the BOCPD framework with the adaptation mechanism for the RL \cite{panda2025curriculum}, to utilize the alternative sensors for fail-safe navigation. 
\printbibliography

@article{liu2022uav,
  title={UAV trajectory optimization for time-constrained data collection in UAV-enabled environmental monitoring systems},
  author={Liu, Kai and Zheng, Jun},
  journal={IEEE Internet of Things Journal},
  volume={9},
  number={23},
  pages={24300--24314},
  year={2022},
  publisher={IEEE}
}

@inproceedings{malhotra2015long,
  title={Long short term memory networks for anomaly detection in time series},
  author={Malhotra, Pankaj and Vig, Lovekesh and Shroff, Gautam and Agarwal, Puneet and others},
  booktitle={Proceedings},
  volume={89},
  number={9},
  pages={94},
  year={2015}
}

@article{psiaki2016gnss,
  title={GNSS spoofing and detection},
  author={Psiaki, Mark L and Humphreys, Todd E},
  journal={Proceedings of the IEEE},
  volume={104},
  number={6},
  pages={1258--1270},
  year={2016},
  publisher={IEEE}
}

@article{montiel2018scikit,
  title={Scikit-multiflow: A multi-output streaming framework},
  author={Montiel, Jacob and Read, Jesse and Bifet, Albert and Abdessalem, Talel},
  journal={Journal of Machine Learning Research},
  volume={19},
  number={72},
  pages={1--5},
  year={2018}
}

@article{tokekar2016sensor,
  title={Sensor planning for a symbiotic UAV and UGV system for precision agriculture},
  author={Tokekar, Pratap and Vander Hook, Joshua and Mulla, David and Isler, Volkan},
  journal={IEEE transactions on robotics},
  volume={32},
  number={6},
  pages={1498--1511},
  year={2016},
  publisher={IEEE}
}

@inproceedings{ruff2018deep,
  title={Deep one-class classification},
  author={Ruff, Lukas and Vandermeulen, Robert and Goernitz, Nico and Deecke, Lucas and Siddiqui, Shoaib Ahmed and Binder, Alexander and M{\"u}ller, Emmanuel and Kloft, Marius},
  booktitle={International conference on machine learning},
  pages={4393--4402},
  year={2018},
  organization={PMLR}
}

@inproceedings{yang2022class,
  title={Class-aware contrastive semi-supervised learning},
  author={Yang, Fan and Wu, Kai and Zhang, Shuyi and Jiang, Guannan and Liu, Yong and Zheng, Feng and Zhang, Wei and Wang, Chengjie and Zeng, Long},
  booktitle={Proceedings of the IEEE/CVF Conference on Computer Vision and Pattern Recognition},
  pages={14421--14430},
  year={2022}
}

@inproceedings{yang2020feedback,
  title={Feedback recurrent autoencoder},
  author={Yang, Yang and Sauti{\`e}re, Guillaume and Ryu, J Jon and Cohen, Taco S},
  booktitle={ICASSP 2020-2020 IEEE International Conference on Acoustics, Speech and Signal Processing (ICASSP)},
  pages={3347--3351},
  year={2020},
  organization={IEEE}
}

@article{li2022unsupervised,
  title={Unsupervised machine anomaly detection using autoencoder and temporal convolutional network},
  author={Li, Zhiyuan and Sun, Yu and Yang, Laihao and Zhao, Zhibin and Chen, Xuefeng},
  journal={IEEE Transactions on Instrumentation and Measurement},
  volume={71},
  pages={1--13},
  year={2022},
  publisher={IEEE}
}

@article{panda2025generative,
  title={Generative Adversarial Evasion and Out-of-Distribution Detection for UAV Cyber-Attacks},
  author={Panda, Deepak Kumar and Guo, Weisi},
  journal={arXiv preprint arXiv:2506.21142},
  year={2025}
}

@article{shao2025efficient,
  title={Efficient Path-Following for Urban Logistics: A Fuzzy Control Strategy for Consumer UAVs under Disturbance Constraints},
  author={Shao, Xingling and Du, Jun and Xia, Yi and Zhang, Zekai and Hou, Xiangwang and Debbah, Merouane},
  journal={IEEE Transactions on Consumer Electronics},
  year={2025},
  publisher={IEEE}
}

@article{adams2007bayesian,
  title={Bayesian online changepoint detection},
  author={Adams, Ryan Prescott and MacKay, David JC},
  journal={arXiv preprint arXiv:0710.3742},
  year={2007}
}

@article{zhang2018intelligent,
  title={Intelligent GNSS/INS integrated navigation system for a commercial UAV flight control system},
  author={Zhang, Guohao and Hsu, Li-Ta},
  journal={Aerospace science and technology},
  volume={80},
  pages={368--380},
  year={2018},
  publisher={Elsevier}
}

@article{panda2025curriculum,
  title={Curriculum-Guided Antifragile Reinforcement Learning for Secure UAV Deconfliction under Observation-Space Attacks},
  author={Panda, Deepak Kumar and Perrusquia, Adolfo and Guo, Weisi},
  journal={arXiv preprint arXiv:2506.21129},
  year={2025}
}

@article{ioannides2016known,
  title={Known vulnerabilities of global navigation satellite systems, status, and potential mitigation techniques},
  author={Ioannides, Rigas Themistoklis and Pany, Thomas and Gibbons, Glen},
  journal={Proceedings of the IEEE},
  volume={104},
  number={6},
  pages={1174--1194},
  year={2016},
  publisher={IEEE}
}

@article{noh2019tractor,
  title={Tractor beam: Safe-hijacking of consumer drones with adaptive GPS spoofing},
  author={Noh, Juhwan and Kwon, Yujin and Son, Yunmok and Shin, Hocheol and Kim, Dohyun and Choi, Jaeyeong and Kim, Yongdae},
  journal={ACM Transactions on Privacy and Security (TOPS)},
  volume={22},
  number={2},
  pages={1--26},
  year={2019},
  publisher={ACM New York, NY, USA}
}

@article{wang2019autonomous,
  title={Autonomous navigation of UAVs in large-scale complex environments: A deep reinforcement learning approach},
  author={Wang, Chao and Wang, Jian and Shen, Yuan and Zhang, Xudong},
  journal={IEEE Transactions on Vehicular Technology},
  volume={68},
  number={3},
  pages={2124--2136},
  year={2019},
  publisher={IEEE}
}

@article{meng2025trusted,
  title={Trusted Multisource Fusion Navigation for UAV Under GNSS Interference and Spoofing Attacks},
  author={Meng, Chen and Hu, Qinglei and Ge, Shuzhi Sam and Li, Dongyu},
  journal={IEEE/ASME Transactions on Mechatronics},
  year={2025},
  publisher={IEEE}
}

@inproceedings{borhani2020deep,
  title={Deep neural network approach to detect GNSS spoofing attacks},
  author={Borhani-Darian, Parisa and Li, Haoqing and Wu, Peng and Closas, Pau},
  booktitle={Proceedings of the 33rd International Technical Meeting of the Satellite Division of The Institute of Navigation (ION GNSS+ 2020)},
  pages={3241--3252},
  year={2020}
}

@article{broumandan2017approach,
  title={An approach to detect GNSS spoofing},
  author={Broumandan, Ali and Siddakatte, Ranjeeth and Lachapelle, G{\'e}rard},
  journal={IEEE Aerospace and Electronic Systems Magazine},
  volume={32},
  number={8},
  pages={64--75},
  year={2017},
  publisher={IEEE}
}

@article{iqbal2024deep,
  title={A deep learning based induced GNSS spoof detection framework},
  author={Iqbal, Asif and Aman, Muhammad Naveed and Sikdar, Biplab},
  journal={IEEE transactions on machine learning in communications and networking},
  year={2024},
  publisher={IEEE}
}

\end{document}